\documentclass[runningheads]{llncs}
\usepackage{graphicx}
\usepackage{amsmath,amssymb,amsfonts}
\usepackage{graphicx}
\usepackage{textcomp}
\usepackage{csquotes}
\usepackage{multirow}
\usepackage[normalem]{ulem}
\usepackage{longtable}
\usepackage{float}
\usepackage{adjustbox}
\begin{document}

\title{Fruit classification using deep feature maps in the presence of deceptive similar classes
}

\author{Mohit Dandekar\inst{1}\orcidID{0000-0002-5104-7496} \and
Narinder Singh Punn\inst{1}\orcidID{0000-0003-1175-1865}
\and
Sanjay Kumar Sonbhadra\inst{1}\orcidID{0000-0002-7457-9655}
\and
Sonali Agarwal\inst{1}\orcidID{0000-0001-9083-5033} 
}
\authorrunning{Mohit et al.}

\institute{ Indian Institute of Information Technology Allahabad, Jhalwa, Prayagraj, Uttar Pradesh, India \\
\email{\{mit2018012, pse2017002, rsi2017502, sonali\}@iiita.ac.in}}
\maketitle              

\begin{abstract}
Autonomous detection and classification of objects are admired area of research in many industrial applications. Though, humans can distinguish objects with high multi-granular similarities very easily; but for the machines, it is a very challenging task. The convolution neural networks (CNN) have illustrated efficient performance in multi-level representations of objects for classification. Conventionally, the existing deep learning models utilize the transformed features generated by the rearmost layer for training and testing. However, it is evident that this does not work well with multi-granular data, especially, in presence of deceptive similar classes (almost similar but different classes). The objective of the present research is to address the challenge of classification of deceptively similar multi-granular objects with an ensemble approach thfat utilizes activations from multiple layers of CNN (deep features). These multi-layer activations are further utilized to build multiple deep decision trees (known as Random forest) for classification of objects with similar appearance. The Fruits-360 dataset is utilized for evaluation of the proposed approach. With extensive trials it was observed that the proposed model outperformed over the conventional deep learning approaches.

\end{abstract}

\section{Introduction}
Understanding, categorizing, and classifying various objects of the world are the most accomplished feats of the human brain. Human visual recognition can learn and remember a diverse set of objects and related features. In a single glance the human eyes can distinguish objects and identify the respective classes. Several machine learning and artificial intelligence algorithms have been developed concerning the same objective to make intelligent machines \cite{lecun2015deep,lu2007survey,punn2020inception,tomar2015twin,alam2020one}. It is evident that recently, the ensemble models \cite{dietterich2000ensemble} are becoming the choice of researchers for pattern recognition, data mining, and image classification tasks~\cite{sarkar2019predictive,rathore2014predicting}. The ensemble model is a method which joins at least two machine learning models seeking for better outcomes. The performance of any machine learning model is measured with respect to bias and variance \cite{rodriguez2006rotation}. The bias portrays effortlessness regarding the model and its higher value indicates under-fitting, whereas how much a model is influenced or harmed by adjustment done on a dataset is identified by the value of variance. More variance shows over-fitting \cite{breiman1996bias}, hence in an ideal model less bias and less variance is expected. Seeking for these objectives, the ensemble methods are established as the best choice for image classification~\cite{sonbhadra2020application}. Therefore, in the present article an ensemble technique is proposed for fruit classification.

Accordning to Zieler et al. \cite{zeiler2014visualizing}, established methods for image classification conventionally use the features of the last fully connected layer for the output layer to make the final decision. The last layer has exceptional responsiveness to semantic knowledge at the category level, whereas the middle layers are least responsive to the semantics, however, it retains a large number of information. It was also observed that the representations from every layer display the structural nature of the network characteristics, where lower layers react to the edges and edge-colour concurrence, while top layers show significantly more class-specific differences.

The present work utilizes bagging \cite{dietterich2000experimental} as an ensemble model that separates the data in numerous subsets, and each subset is utilized separately by independent models for training. Later, the output of each model is collected to decide the end outcome. This work also utilizes random forest (RF) classifier \cite{liaw2002classification} aimed at class identification. Decision trees suffer with greater variance thus prone to over-fit \cite{safavian1991survey} whereas, RF utilizes many decision trees to minimize variance and bias that makes this classifier a better choice \cite{rodriguez2006rotation}. 

The present research work is aimed to create a model for classifying deceptively similar classes i.e. it can distinguish objects belonging to different categories as well as it also needs to classify the objects of the same category. CNNs have shown remarkable ability in multi-level illustration of an object \cite{gu2018recent}. The disadvantage of the recent deep learning approaches \cite{sa2016deepfruits,hemming2014fruit,dai2016r,murecsan2018fruit} for classification is that they only use the features of the last layer for the training of flat \textit{N}-way classifier. This paper proposes a novel approach that utilises the features of the multiple level abstraction of the convolution layers along with the dense layer for the construction of multiple deep decision trees (forest) that can identify deceptively similar objects (common appearances) effectively. From experimental results it is observed that for multi-granular data, the forest of deep features can enhance the classification accuracy with multiple levels of abstraction.

The rest of the paper is organized as follows: Section 2 briefs the recent research contributions whereas the proposed model is discussed in section 3. Details of the experimental setup are discussed in section 4 whereas the results are discussed in section 5 followed by concluding remarks and future scope in the last section.

\section{Related Work}
The deep neural networks have made substantial improvements in pattern recognition and classification tasks~\cite{krizhevsky2012imagenet,szegedy2015going,simonyan2014very}. One such promising task is fruit classification where a variety of factors like fluctuating brightness scenes, occlusion, sharp edges, shapes, reflective properties, etc. make the process challenging. Presence of deceptively similar objects makes the classification task more difficult, and to solve this problem very limited research articles have been reported till date. Zong et al.~\cite{ge2015modelling} have utilized Gaussian mixture model in classification of deceptive similar classes, and afterwards concerning the same Zheng et al.~\cite{zheng2017learning} used multi-attention convolution neural network for classifying birds, aircraft and cars. 

From the context of fruit classification, many research works have been done with the help of deep learning approaches. Sa et al.~\cite{sa2016deepfruits} proposed a faster region-based CNN with transfer learning, where RGB and NIR (Near-Infrared) pictures  were utilized to train the model. The integration of RGB and NIR uncover the techniques of early and late fusion methods. In addition, identification in relation to camera angle~\cite{hemming2014fruit}, scale invariant feature transform (SIFT), enhanced ChanVese level-set model~\cite{chan2001active} are often utilised for fruit recognition tasks. Faster R-CNN architecture~\cite{bargoti2017deep} is another method that illustrates state-of-the-art techniques for fruit identification effectively in orchards, including mangoes, almonds, and apple. Another approach to faster region-based CNN has been utilized to automatically harvest and detect fruit from images, here the network is equipped using RGB and NIR images to achieve improved efficiency~\cite{dai2016r}. It is also observed that for automated harvesting and farming several deep learning approaches have been proposed in recent years~\cite{harrell1989fruit,sa2016deepfruits}. Concerning the same; Specially, several CNN architectures have been utilized for identification and labeling of fruits/vegetables~\cite{wu2016personal,sunderhauf2014fine,murecsan2018fruit}. In the research work~\cite{murecsan2018fruit}, the CNN structure consists of several convolution layers, max-pooling layers with strides, and fully connected (dense) layers. To improve the accuracy, Grayscale images were produced to increase the overall channel depth from 3 (RGB) to 4 (RGB + Grayscale) followed by pre-processing to optimize the results. CNN was trained for more than 40,000 iterations for a batch size of 50 images for identifying fruit images and achieved  96.3\% accuracy. 

It is observed from above discussed state-of-the-art methods that none of the researches have considered the classification related problems raised by the presence of deceptive similar classes. Motivated by above discussed notions, the present work proposes a novel technique for fruit classification using standard CNN architecture along with transfer learning and deep random forest classifiers, in presence of deceptive similar classes. 

\section{Proposed approach}
Multiple convolution neural networks are used in this research which comprises a combination of various layers like convolution, max-pooling, rectified linear unit (ReLu) activation, dense and loss. In a standard CNN architecture, each convolution layer is accompanied with a ReLu activation layer followed by a pooling layer and finally several fully connected dense layers connected to the decision layer~\cite{zeiler2014visualizing}. Attribute which distinguishes the CNN from a normal neural network is that it considers the image composition while processing it. 

\begin{figure}
  \includegraphics[width=\textwidth]{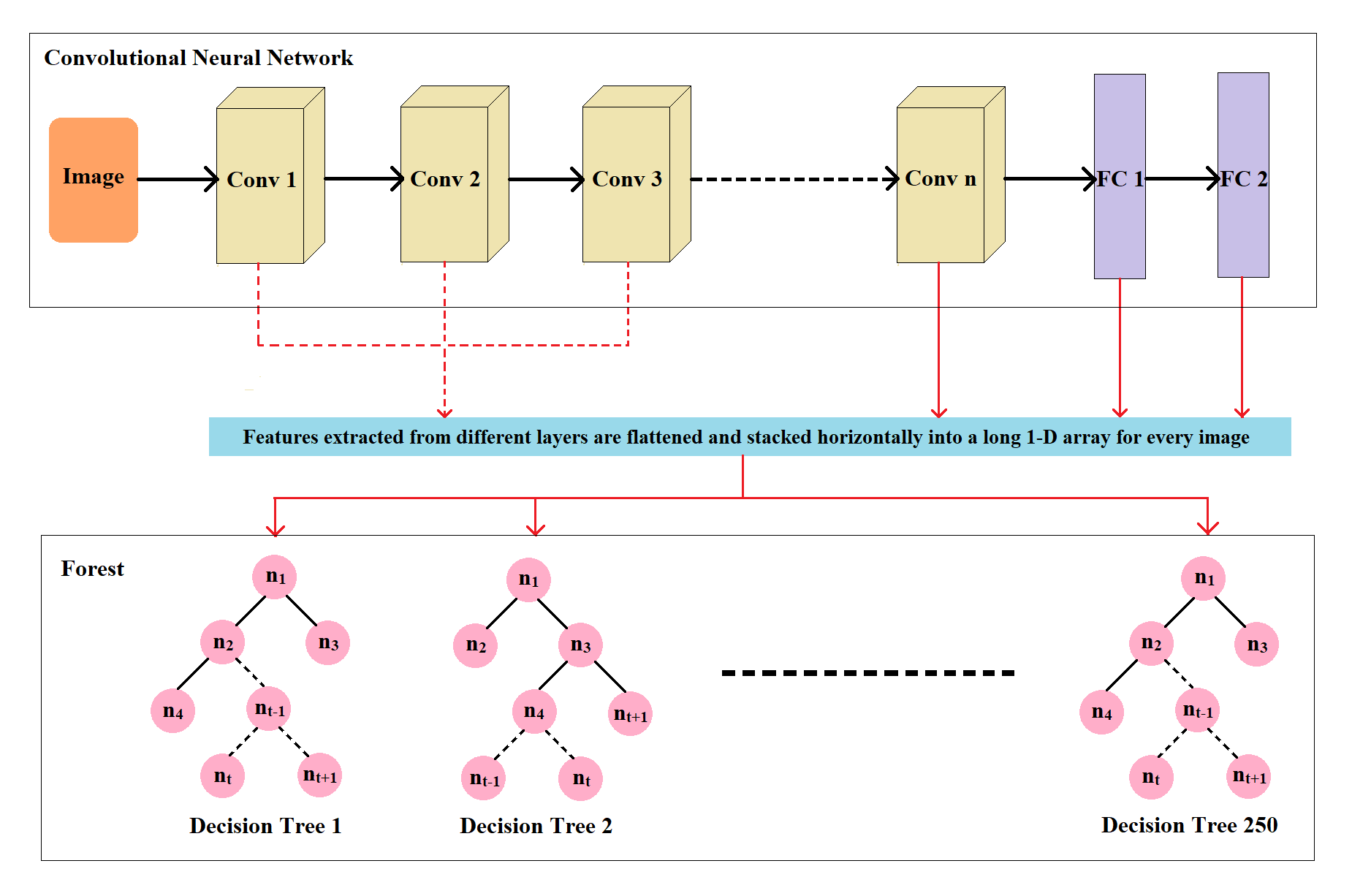}
\caption{The proposed framework.}
\label{fig:1}
\end{figure}

First, the CNN models are trained on Fruit-360 dataset~\cite{murecsan2018fruit} until the model fails to improve the validation score using the earlystopping technique. Then, the features from multiple layers are taken out from the CNN network as the input for formulating multiple decision trees. For every image, the layer outputs are flattened and stacked horizontally, thereby forming 1-D array for every image. Using these 1-D arrays, multiple trees are learned using bagging and random forest classifier technique to create the ensemble model. The schematic representation of the overall approach is presented in Fig.~\ref{fig:1}.

\subsection{Dataset description}

The Fruits-360 dataset~\cite{murecsan2018fruit} contains 81120 images of 120 categories of fruits, where 75\% samples are used for training and remaining 25\% for testing. The pictures were acquired by filming fruits while being revolved by a motor while using a white sheet of paper as the backdrop and a Flood fill [citation] type algorithm for extracting the fruit from its background. The images are labelled by marking each pixel along with its neighbouring pixel for which colour gap is less than certain threshold. This has been replicated until pixels can no longer be labeled. The marked pixels are assumed to be backdrop (that is later filled with white colour), and the remaining pixels represent the fruit. The largest distance value between 2 adjacent pixels is an algorithm parameter, which is defined (by empirical observations) for every video. The fruits were calibrated to fit into 100 $\times$ 100 pixels image.
\begin{figure}
  \includegraphics[width=\textwidth]{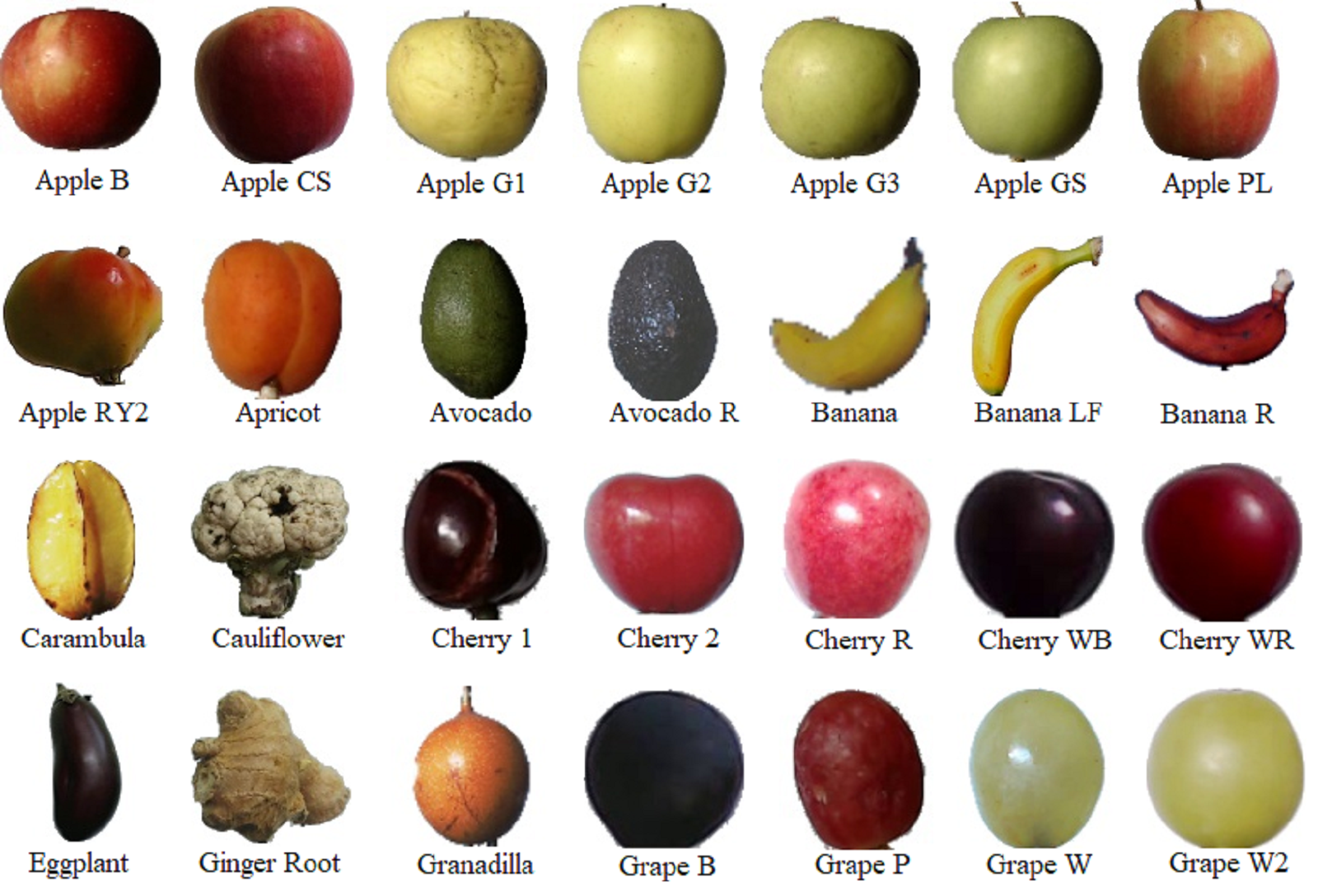}
\caption{Illustration of few images from Fruits-360 dataset.}
\label{fig:2}
\end{figure}

\subsection{Objective function}
The paper uses categorical cross-entropy loss which works on the concepts of softmax loss function. This has a softmax activation including a cross-entropy loss. The softmax loss is given in Eq.~\ref{eq1} and cross-entropy loss is given in Eq.~\ref{eq2}.

\begin{equation}
f\left(s\right)_i=\frac{e^{s_i}}{\sum_{j}^{C}e^{s_j}}
\label{eq1}
\end{equation}
\begin{equation}
CE=-\sum_{i}^{C}{t_ilog}\left(s_i\right)
\label{eq2}
\end{equation}
where \textit{C} is the total number of classes, \textit{s} is a vector comprising of the C output neurons of the CNN and on each $s_i$ the softmax function cannot be exercised independently, as it relies on every element of \textit{s}.
In Eq.~\ref{eq2}, $t_i$ and $s_i$ are the true value and the CNN output for each $class_i$ in \textit{C}.

The categorical cross-entropy loss given in Eq.~\ref{eq3} for backpropagation in the CNN models, compares the predicted distribution (the output from the softmax layer; one for each class) with the original distribution (where the probability of true class is \enquote*{1} and other classes are \enquote*{0}). In other words, the true class is represented as a one-hot encoded vector, so the nearer the output of the model is to this vector, the less is the loss.
\begin{equation}
CE=-\sum_{i}^{C}{t_ilog\left(f\left(s\right)_i\right)}
\label{eq3}
\end{equation}
Since $t_i$ has only two values \enquote*{0} or \enquote*{1}, the vector t has only one element which is not zero i.e. $t_i=t_p$. So, Eq.~\ref{eq3} can be rewritten as in Eq.~\ref{eq4}.
\begin{equation}
CE=-log\left(\frac{e^{s_p}}{\sum_{j}^{C}e^{s_j}}\right)
\label{eq4}
\end{equation}
here $s_p$ is CNN output belonging to the positive class.
The gradient of the loss function for the positive and negative class is given in Eq.~\ref{eq5} and Eq.~\ref{eq6} respectively. These are used to update the trainable parameters and reduce the loss in subsequent epochs.
\begin{equation}
\frac{\partial}{\partial s_p}\left(-log\left(\frac{e^{s_p}}{\sum_{j}^{C}e^{s_j}}\right)\right)=\left(\frac{e^{s_p}}{\sum_{j}^{C}e^{s_j}}-1\right)
\label{eq5}
\end{equation}
\begin{equation}
\frac{\partial}{\partial s_n}\left(-log\left(\frac{e^{s_p}}{\sum_{j}^{C}e^{s_j}}\right)\right)=\left(\frac{e^{s_n}}{\sum_{j}^{C}e^{s_j}}\right)
\label{eq6}
\end{equation}
Following are the attributes of the forest of multiple deep decision trees:
\begin{itemize}
\item The number of deep decision trees in the forest equals 250.
\item Gini impurity is the feature used to calculate the quality of a split.
\item The number of attributes to evaluate when searching for the best split is equal to the square root of the total number of features.
\end{itemize}

\subsection{Hyperparameters of the model}
Hyperparameters are the variables that defines the network structure and needs to be tuned prior to the training to efficiently learn the deep hidden patterns associated with the desired object or any task. Following are the hyperparamters that needs to be adjusted for efficient training of the deep neural network models: epoch, batch size, dropout, filter size, number of decision trees, quality of split, padding, stride, number of attributes to evaluate, etc.

\section{Experimentation \& Results}
\subsection{Training and Testing}
The input shape is used as 100 $\times$ 100 $\times$ 4 for all the CNN models utilized in this work. The paper implements a custom layer that transforms the original image from RGB to HSV and grayscale, and concatenates the results forming an input of shape 100 $\times$ 100 $\times$ 4. Also, Adadelta optimizer~\cite{zeiler2012adadelta} is used  with a learning rate of 0.1 which reduces on a plateau with learning rate reduction factor of 0.5 and patience value as 3 which is the number of epochs to wait before reducing the learning rate when the loss plateaus. The metrics used for evaluation of the model is classification accuracy, precision, recall, specificity and F1-score represented in the following equations.
\begin{equation}
Accuracy=\frac{TP+TN}{TP+TN+FP+FN}
\label{eq7}
\end{equation}
\begin{equation}
Precision=\frac{TP}{TP+FP}
\label{eq8}
\end{equation}
\begin{equation}
Recall=\frac{TP}{TP+FN}
\label{eq8}
\end{equation}
\begin{equation}
Specificity=\frac{TN}{TN+FP}
\label{eq8}
\end{equation}
\begin{equation}
F1score=\frac{2\times Recall\times Precision}{Recall + Precision}
\label{eq8}
\end{equation}
where true positive (TP) is equivalent to positive prediction, true negative (TN) indicates correct rejection, false positive (FP) highlights incorrect positive prediction, and false negative (FN) shows miss classification.
The optimizer used for training the CNNs is Adadelta with the running average $E\left[g^2\right]_t$ given in Eq.~\ref{eq8} along with the update rule for this optimizer is given in Eq.~\ref{eq9}.
\begin{equation}
E\left[g^2\right]_t=\gamma E\left[g^2\right]_{t-1}+\left(1-\gamma\right)g_t^2	
\label{eq8}
\end{equation}
\begin{equation}
\Theta_{t+1}=\Theta_t+\Delta\Theta_t
\end{equation}
\begin{equation}
\Delta\Theta_t=-\frac{\eta}{\sqrt{E\left[g^2\right]_t+\epsilon}}g_t
\label{eq9}
\end{equation}
where $\eta$ is the learning rate, $g_t$ is the gradient at time step \textit{t},  $\Theta$ is the model's parameter
and $\gamma$ is the momentum term.

The feature used to calculate the quality of split for training the decision trees is Gini impurity as represented in Eq.~\ref{eq11}.
\begin{equation}
G=\sum_{i=1}^{C}p\left(i\right)\cdot\left(1-p\left(i\right)\right)
\label{eq11}
\end{equation}
where \textit{C} is total number of classes and \textit{p(i)} is the probability of  datapoint selection with class \textit{i}.

\subsubsection{Experiment 1: 4-layer CNN Model}
This model has been proposed by Murescan et al.~\cite{murecsan2018fruit} for fruit classification using Fruits-360 dataset. In present research, initially, the above discussed model is trained, and then the features are extracted from the last convolution layer and the two fully connected dense layers to create a forest of deep decision trees. Table~\ref{tab:1} shows the architecture of the 4-layer CNN model.
\begin{table}
\caption{The 4-layer CNN Architecture}
\centering
\label{tab:1}
\begin{tabular}{ccccc}
\hline\noalign{\smallskip}
Layer type & Filter used & Output Shape & Activation & Params.\\
 & Size/Strides/Padding & & & \\
\noalign{\smallskip}\hline\noalign{\smallskip}
Lambda (input) & - & 100 x 100 x 4 & - & 0 \\
convolution 2D & 5x5 / 1x1 / same & 100 x 100 x 16 & ReLu & 1616 \\
Max pooling 2D & 2x2 / 2x2 / valid & 50 x 50 x 16 & - & 0 \\
convolution 2D & 5x5 / 1x1 / same & 50 x 50 x 32 & ReLu & 12832 \\
Max pooling 2D & 2x2 / 2x2 / valid & 25 x 25 x 32 & - & 0 \\
convolution 2D & 5x5 / 1x1 / same & 25 x 25 x 64 & ReLu & 51264 \\
Max pooling 2D & 2x2 / 2x2 / valid & 12 x 12 x 64 & -- & 0 \\
convolution 2D & 5x5 / 1x1 / same & 12 x 12 x 128 & ReLu & 204928 \\
Max pooling 2D & 2x2 / 2x2 / valid & 6 x 6 x 128 & - & 0 \\
Dense & - & 1024 & ReLu & 4719616 \\
Dense & - & 256 & ReLu & 131200 \\
Dense & - & 120 & Softmax & 15480 \\
\noalign{\smallskip}\hline
\end{tabular}
\end{table}
\begin{figure}
  \includegraphics[width=\textwidth]{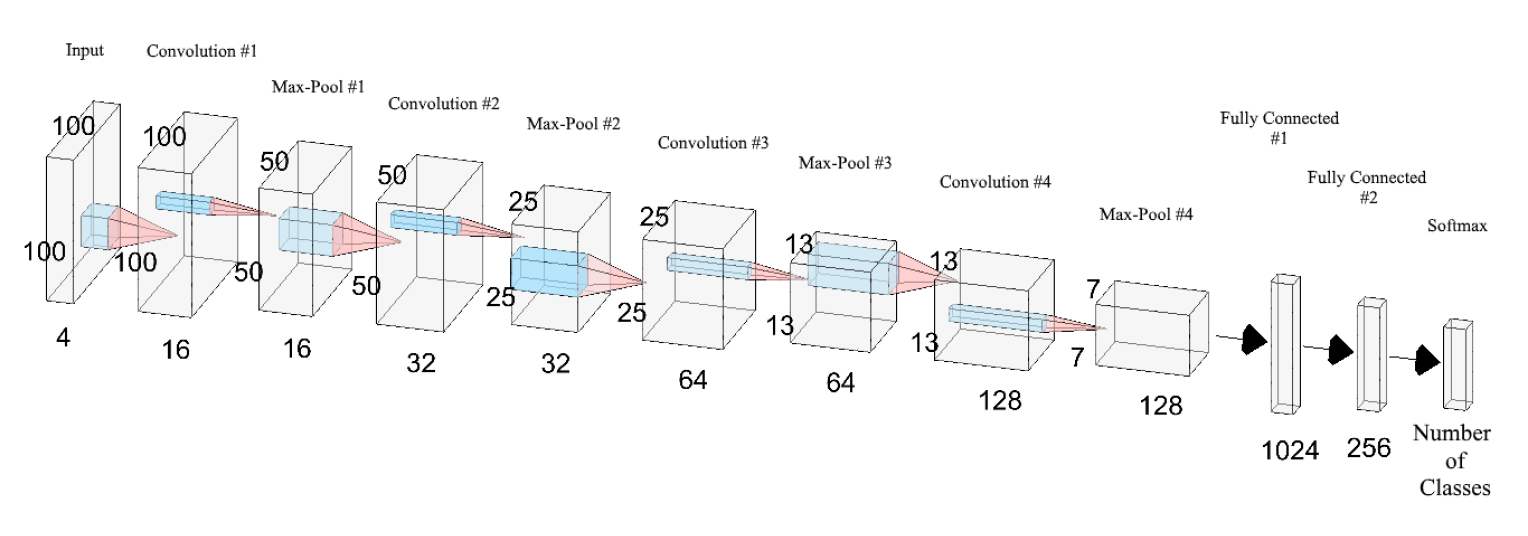}
\caption{Visualization of 4-layer CNN model~\cite{murecsan2018fruit}}
\label{fig:3}
\end{figure}

\subsubsection{Experiment 2: VGG16 Architecture}
\label{experiment 2}
\begin{figure}
  \includegraphics[width=\textwidth]{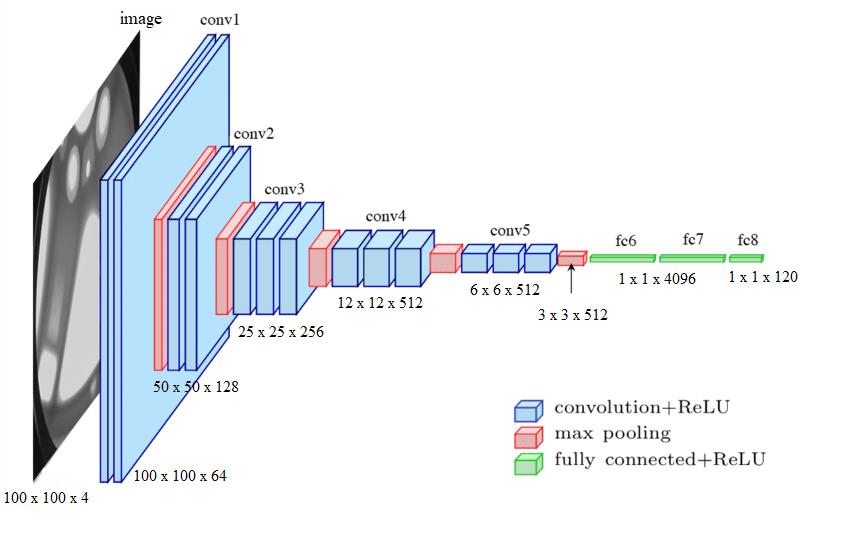}
\caption{Visualization of the VGG architecture \cite{vgg16arch}}
\label{fig:4}
\end{figure}

VGG16 is a model proposed by Simonyan and Zisserman from the University of Oxford \cite{simonyan2014very} as shown by Fig.~\ref{fig:4}. In ImageNet \cite{deng2009imagenet}, which is a dataset of over 14 million images belonging to 1000 groups, the model attains 92.7\% top-5 test accuracy. This is one of the renowned models presented in ILSVRC 2014 \cite{russakovsky2015imagenet}. It improves on AlexNet \cite{zeiler2014visualizing} by substituting heavy kernel-sized filters (11 and 5 respectively in the initial two convolution layers) with numerous 3 $\times$ 3 kernel-sized filters in succession.

In this research the VGG16 architecture has been modified as follows:
\begin{enumerate}
\item Custom input, 100 $\times$ 100 $\times$ 4 pre-processed images.
\item 2 fully-connected (dense) layers with 4096 neurons with a ReLu activation function.
\item Softmax classifier with 120 neurons for the number of classes of fruits.
\item Adadelta optimizer with a reducing learning rate.
\end{enumerate}

After the above discussed modifications the VGG16 model \cite{simonyan2014very} is used with pre-trained weights of the ImageNet \cite{deng2009imagenet} dataset. Afterwards, all the layers are made trainable and then the modified model is trained on the Fruits-360 dataset \cite{murecsan2018fruit}. Finally, the features are extracted from the last convolution layer and the two fully connected dense layers for creating a forest of deep decision trees.

\subsubsection{Experiment 3: ResNet50 architecture}

\begin{figure}
  \includegraphics[width=\textwidth]{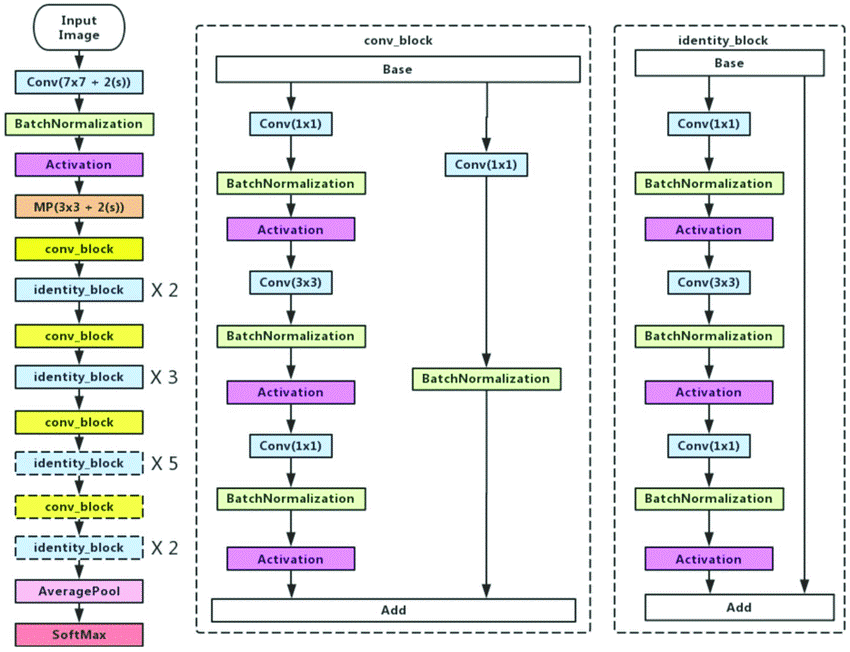}
\caption{(Left) Visualization of ResNet50 architecture. (Centre) Convolution block that alters the input shape. (Right) Identity block that does not alter the input shape. \cite{ji2019optimized}}
\label{fig:5}
\end{figure}

ResNet-50 \cite{he2016deep} is a deep residual network consisting of \enquote{50} layers comprising of combination of convolution and pooling layers. Its key breakthrough is the skip connection. Deep networks often suffer from vanishing gradients without adjustments, meaning the gradient becomes smaller as the model backpropagates. The skip link enables the network to learn identity feature, which allows it to move the input through the block, without going through the other weight layers. This allows additional layers to be stacked and a deeper network to be created, offsetting the vanishing gradient by allowing the network to skip through layers. ResNet is also the winner of ILSVRC 2015 \cite{russakovsky2015imagenet} in image classification, detection and localization, as well as winner of MS COCO 2015 \cite{lin2014microsoft} detection, and segmentation. Fig.~\ref{fig:5} shows the overview of the ResNet-50 architecture.

To make ResNet-50 workable for present classification task in this paper, following modifications have been performed:
\begin{enumerate}
    \item Custom input, 100 $\times$ 100 $\times$ 4 pre-processed images.
    \item 1 fully-connected (dense) layer with 512 neurons with ReLu activation function after the average pooling layer.
    \item Softmax classifier with 120 neurons for the number of classes of fruits.
    \item Adadelta optimizer with a reducing learning rate.
\end{enumerate}

The ResNet-50 model~\cite{he2016deep} is used which is already trained on ImageNet \cite{deng2009imagenet} dataset and making the above stated modifications. After this, all the layers are made trainable and then the modified model is trained on the Fruits-360 dataset \cite{murecsan2018fruit}. Finally, the features are extracted from the last convolution layer, the average pooling layer, and the dense layer for creating a forest of deep decision trees.

\subsection{Results and discussion}

The Table~\ref{tab:2} presents the classification performance of different models with and without using the proposed ensemble technique with the help of accuracy. Among the discussed architectures, the proposed methodology outperformed the performance of standalone standard CNN architectures, whereas it is also observed that ResNet50 preformed better than other models as highlighted in Table~\ref{tab:2}. The proposed approach is also evaluated to classify deceptive similar classes of category apple, cherry, grape, pear and tomato as shown in Table~\ref{tab:3}. Each of these class categories are evaluated using the discussed architectures and proposed ensemble approach as presented in Table~\ref{tab:4}.

\begin{table}
\centering
\begin{center}

\caption{Classification accuracy of fruit images using different approaches.}
\label{tab:2}
\begin{tabular}{|p{1.8cm}|p{2.5cm}|p{3cm}|p{3cm}|}
\hline
Model & Validation accuracy (\%) & Test accuracy without ensemble technique (\%) & Test accuracy (\%) using  the proposed architecture \\
\hline
4-layer CNN & 99.0365 & 97.1244 & 98.3222 \\
VGG16 & 99.8671 & 99.2338 & 99.3114\\
ResNet50 & 99.9336 & 99.8157 & 99.8302\\ \hline
\end{tabular}
\end{center}
\end{table}

\begin{table}
\centering
\caption{Samples of classes with their subcategories available in Fruit-360 dataset.}
\label{tab:3}
\begin{tabular}{|p{1.5cm}|p{10cm}|}
\hline
Classes &     Sub-classes \\ \hline
Apple & Apple Braeburn, Apple Crimson Snow, Apple Golden 1, Apple Golden 2, Apple Golden 3, Apple Granny Smith, Apple Pink Lady, Apple Red 1, Apple Red 2, Apple Red 3, Apple Red Delicious, Apple Red Yellow 1, Apple Red Yellow 2 \\ \hline
Cherry & Cherry 1, Cherry 2, Cherry Rainier, Cherry Wax Black, Cherry Wax Red, Cherry Wax Yellow\\ \hline
Grape & Grape Blue, Grape Pink, Grape White, Grape White 2, Grape White 3, Grape White 4\\ \hline
Pear & Pear, Pear Abate, Pear Forelle, Pear Kaiser, Pear Monster, Pear Red, Pear Williams\\ \hline
Tomato & Tomato 1, Tomato 2, Tomato 3, Tomato 4,
 Tomato Cherry Red, Tomato Maroon, Tomato Yellow\\
\hline
\end{tabular}
\end{table}

 The best performance is given by the ResNet50 model with accuracy, precision, recall, specificity and F1 score. These results imply that using the stated architectures along with the ensemble approach the performance gets improved. The observed improvement is due to the use of a non-differentiable module for predictions rather than using the fully connected dense neural network.

\begin{table}
\caption{Average performance comparison of proposed approach.}
\label{tab:4}
\centering
\resizebox{\textwidth}{!}{
\begin{tabular}{|p{2cm}|c|p{1cm}|p{1.3cm}|p{1.3cm}|p{1.3cm}|p{1.3cm}|p{1.5cm}|}
\hline
Ensemble Model                & Category & No. of classes & Accuracy (Avg.) & Precision (Avg.) & Recall (Avg.)& F1-score (Avg.) & Specificity (Avg.) \\ \hline
\multirow{5}{*}{4- layer CNN} & Apple    & 13             & 0.9994   & 0.9789    & 0.9544 & 0.9648   & 0.9998      \\ \cline{2-8} 
                              & Cherry   & 6              & 0.9998   & 1.0       & 0.9878 & 0.9936   & 1.0         \\ \cline{2-8} 
                              & Grape    & 6              & 0.9999   & 0.9921    & 1.0    & 0.9960   & 0.9999      \\ \cline{2-8} 
                              & Pear     & 7              & 0.9995   & 0.97429   & 0.9829 & 0.9779   & 0.9997      \\ \cline{2-8} 
                              & Tomato   & 7              & 0.9999   & 0.9988    & 1.0    & 0.9994   & 0.9999      \\ \hline
\multirow{5}{*}{VGG16}        & Apple    & 13             & 0.9992   & 0.9399    & 0.9618 & 0.9488   & 0.9994      \\ \cline{2-8} 
                              & Cherry   & 6              & 0.9997   & 0.9796    & 0.9959 & 0.9874   & 0.9997      \\ \cline{2-8} 
                              & Grape    & 6              & 0.9999   & 0.9940    & 1.0    & 0.9970   & 0.9999      \\ \cline{2-8} 
                              & Pear     & 7              & 0.9996   & 0.9947    & 0.9622 & 0.9770   & 0.9999      \\ \cline{2-8} 
                              & Tomato   & 7              & 0.9988   & 0.9503    & 0.9552 & 0.9467   & 0.9993      \\ \hline
\multirow{5}{*}{ResNet50}     & Apple    & 13             & 0.9999   & 1.0       & 0.9995 & 0.9997   & 1.0         \\ \cline{2-8} 
                              & Cherry   & 6              & 0.9999   & 0.9922    & 1.0    & 0.9960   & 0.9999      \\ \cline{2-8} 
                              & Grape    & 6              & 1.0      & 1.0       & 1.0    & 1.0      & 1.0         \\ \cline{2-8} 
                              & Pear     & 7              & 1.0      & 1.0       & 1.0    & 1.0      & 1.0         \\ \cline{2-8} 
                              & Tomato   & 7              & 1.0      & 1.0       & 1.0    & 1.0      & 1.0         \\ \hline
\end{tabular}}
\end{table}

\section{Conclusion}
The deep learning models are usually built using neural networks, i.e., several layers of parameterized nonlinear differentiable modules which can be learned by backpropagation. The proposed approach uses different CNN layers' outputs along with fully connected (dense) layers for creating multiple deep decision trees (forest) to classify deceptively similar multi-granular fruits. With the extensive trials performed using baseline CNN, VGG16 and ResNet50 architectures over Fruit-360 dataset, it is observed that the proposed deep random forest ensemble classification approach performed better as compared to the raw architectures; furthermore, among the ensembled architectures the ResNet50 outperformed others in terms of accuracy, precision, specificity, recall and F1-score. It is also observed that in the presence of deceptive similar classes the proposed ensemble approach was able to distinguish similar fruits better than the conventional models. It is believed that this approach can be further explored to other application domains specially, for anomaly and novelty detection tasks.


%
%


\bibliographystyle{spmpsci}
\bibliography{bibliography.bib}

\end{document}